\pdfoutput=1

\documentclass[11pt, table]{article}

\usepackage[final]{acl}

\usepackage{times}
\usepackage{latexsym}

\usepackage[T1]{fontenc}

\usepackage[utf8]{inputenc}

\usepackage{microtype}

\usepackage{inconsolata}

\usepackage{graphicx}
\usepackage{booktabs}
\usepackage{tabularx}
\usepackage{longtable}
\usepackage{subcaption}
\usepackage{float}

\title{IBSEN: Director-Actor Agent Collaboration for Controllable and Interactive Drama Script Generation}

\author{Senyu Han$^1$, Lu Chen$^{1,3}$\thanks{Corresponding author}, Li-Min Lin$^2$, Zhengshan Xu$^2$, Kai Yu$^{1,3}$ \\
        \textsuperscript{1}X-LANCE Lab, Department of Computer Science and Engineering \\
        MoE Key Lab of Artificial Intelligence, SJTU AI Institute \\
Shanghai Jiao Tong University, Shanghai, China\\ 
        \textsuperscript{2}Department of Cultural Industry Management, School of Media and Communication \\
        Shanghai Jiao Tong University, Shanghai, China \\
        \textsuperscript{3}Suzhou Laboratory, Suzhou, China \\
        \texttt{\{cnlnpjhsy, chenlusz\}@sjtu.edu.cn} \\
        }

\begin{document}
\maketitle
\begin{abstract}
Large language models have demonstrated their capabilities in storyline creation and human-like character role-playing. Current language model agents mainly focus on reasonable behaviors from the level of individuals, and their behaviors might be hard to constraint on the level of the whole storyline. In this paper we introduce IBSEN, a director-actor coordinate agent framework that generates drama scripts and makes the plot played by agents more controllable. The director agent writes plot outlines that the user desires to see, instructs the actor agents to role-play their characters, and reschedules the plot when human players participate in the scenario to ensure the plot is progressing towards the objective. To evaluate the framework, we create a novel drama plot that involves several actor agents and check the interactions between them under the instruction of the director agent. Evaluation results show that our framework could generate complete, diverse drama scripts from only a rough outline of plot objectives, meanwhile maintaining the characteristics of characters in the drama. Our codes and prompts are available at \url{https://github.com/OpenDFM/ibsen}.
\end{abstract}

\section{Introduction}

Language models learn the commonsense knowledge and reasoning ability from pre-trained text data, and they are suitable for many generation tasks such as role-playing \cite{wang2023rolellm} and story creation \cite{see2019massively, wang2023open}. Utilizing the knowledge stored in model parameters, using language models for story creation can rapidly and efficiently generate diverse storylines, potentially inspiring creators with new ideas. Recently, the rapid progress of large language models (LLMs) greatly improved their reasoning ability, and capable LLMs like ChatGPT\footnote{\url{https://chat.openai.com/}} and GPT-4 \cite{openai2023gpt42} could be used to simulate human intelligence. This not only provides creators with more powerful story creation abilities via language models, but also enables LLM to vividly role-play the characters in the story. A milestone to simulate believable human behavior using LLM is Generative Agents \cite{park2023generative}. It created a virtual world where LLM-based characters make their day and interact with each other. By combining LLM's ability of storyline generation and agent role-playing, creators can easily bring a vivid story setting to life.

\begin{figure}[t]
    \centering
    \includegraphics[width=\columnwidth]{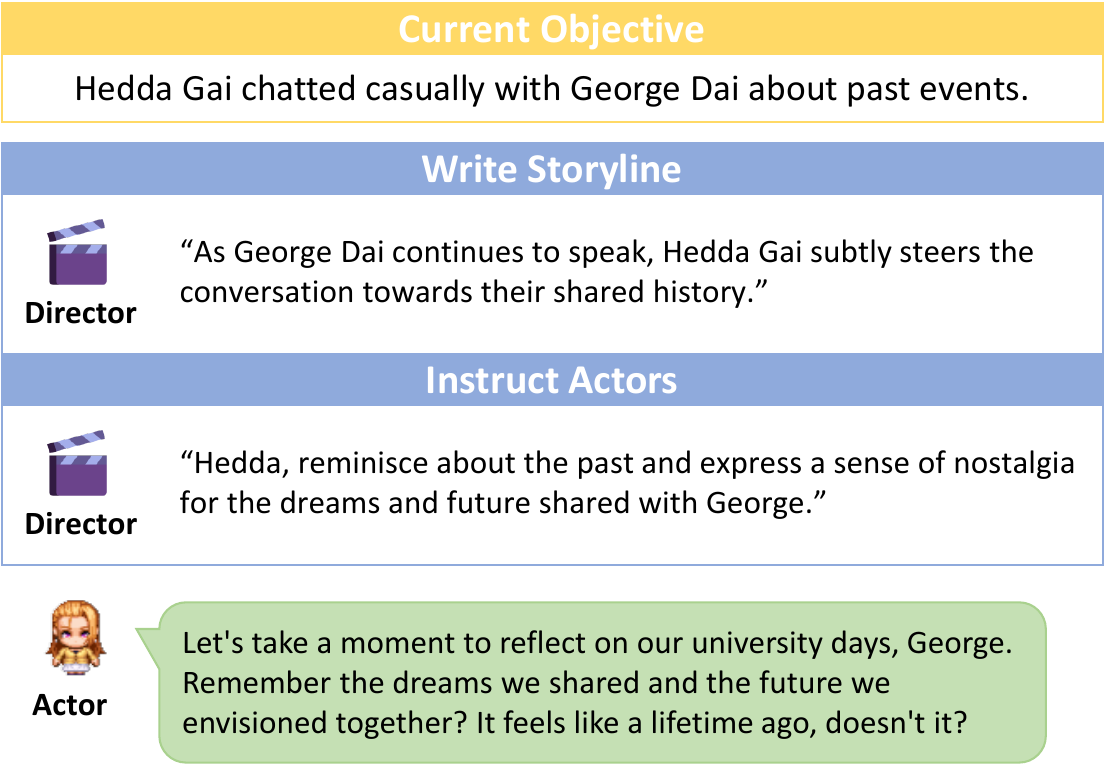}
    \caption{IBSEN is an agent framework that uses director and actor agents to generate controlled drama scripts.}
    \label{fig:introfig}
    \vspace{-4mm}
\end{figure}

However, as the previous research on LLM agents concentrates on the framework of an individual agent, the multi-agent interactions could be greatly diverse and freestyle. The events between different characters are entirely based on their natural interactions, rather than a purposeful storyline. In some cases like drama playing, characters are expected to make certain decisions in a series of scenarios to normally push the plot development forward, and the interactions between agents should be constrained. Only improving the architecture of agents hardly solves this problem, as the distributed agents lack a centralized component to regulate their behaviors. In other cases like adventure games or script role-playing games, the involvement of human players would make the plot development more unpredictable, and it is more necessary to ensure the plot develops towards the correct direction and will not deviate from the original plot too much. Currently, the control of the generation process is mainly focused on controlling individual storyline LLMs \cite{prabhumoye-etal-2020-exploring, zhang2023survey}, while the work on controlling generation in the form of character agents is still lacking.

\begin{figure*}[t]
    \centering
    \includegraphics[width=\textwidth]{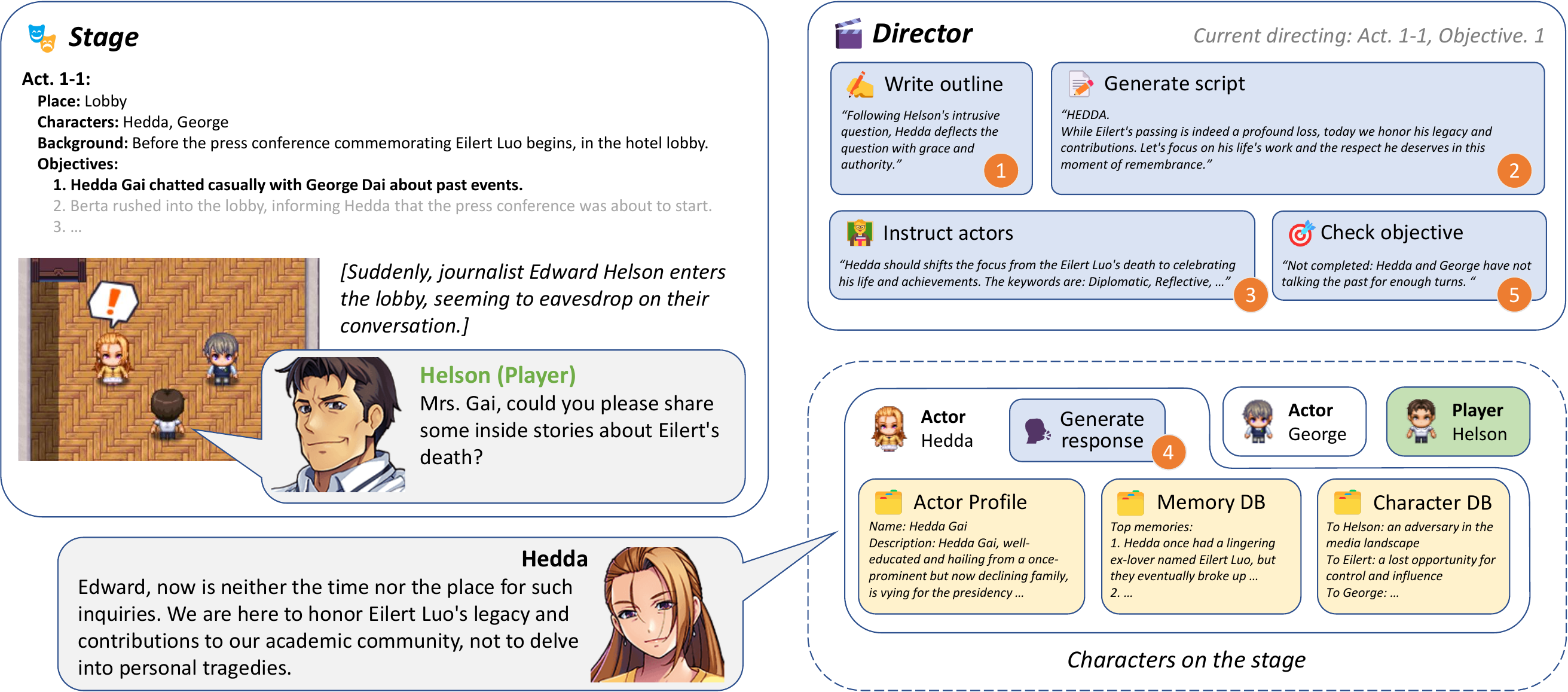}
    \caption{Overall framework of IBSEN and an illustration of director agent's controlling process in a certain scene. Director agent \raisebox{.5pt}{\textcircled{\raisebox{-.9pt} {1}}} writes an outline to continue the previous plot and \raisebox{.5pt}{\textcircled{\raisebox{-.9pt} {2}}} translate it into the dialogue script, then \raisebox{.5pt}{\textcircled{\raisebox{-.9pt} {3}}} instructs the actor agent to generate a proper response. Actor agent in the next dialogue turn \raisebox{.5pt}{\textcircled{\raisebox{-.9pt} {4}}} generates the response, and the director agent \raisebox{.5pt}{\textcircled{\raisebox{-.9pt} {5}}} checks whether the dialogue has reached the current plot objective.}
    \label{fig:overall}
    \vspace{-2mm}
\end{figure*}

To realize both storyline generation and multi-agent collaboration in the drama playing field, we propose the framework of \textbf{IBSEN} (\textbf{I}nteractive-\textbf{B}ased playhou\textbf{S}e for ag\textbf{EN}ts). IBSEN focuses on the controllable generation of drama scripts in the form of agent dialogues. Inspired by the roles of director and actor in films and theatres, we propose the director agent to serve as that centralized component. While character role-playing agents called actors have the freedom to generate responses, they are instructed by the director to make their responses follow the plot development (Figure \ref{fig:introfig}). The implementation of the IBSEN framework is largely prompt-based, and one can easily construct IBSEN agents on any publicly available general LLMs without fine-tuning. Although we test and evaluate IBSEN in a drama scenario, its framework design could be expanded to broader application cases, especially agent-based interactive games that contain specialized plot objectives.

Our contributions are summarized as follows: 
\begin{itemize}
\setlength\itemsep{-0.3em}
    \item We propose the IBSEN, an LLM-based framework for generating controllable drama scripts with agent characters.
    \item Our proposed framework allows human players to participate in the plot and dynamically adjust plot details according to player actions.
    \item We conduct preliminary quantitative and qualitative analysis of the framework's performance, verifying its effectiveness in generating drama scripts.
\end{itemize}

\section{Related Work}
 
\noindent\textbf{Story-telling Using Language Models}\quad Previous works have attempted to create stories and control the generated plots using language models. Users could use plain text to prompt the model, or inject story knowledge into model parameters to control the generation \cite{wang2023open}.  For better story coherency, \citet{yao2019plan} proposed a hierarchical story generation strategy to maintain the logic of the whole storyline. \citet{mirowski2023co} used language models to build an interactive drama script co-writing tool, and creators could directly interfere with the plot generation by prompting the model. As storylines are largely driven by the characters, controlling character scripts is an effective method to control the plot development \cite{dirik2021controlled, xu2020controllable}. From this point, IBSEN mainly uses agent-style characters to automatically build personalized drama dialogues and push the plot forward.

\noindent\textbf{Role-playing Using Language Models}\quad Language models require character-related knowledge to output personalized responses. If the character is derived from real-world or fictional work, its documents and profiles could be collected to serve as a reference to the language model \cite{li2023chatharuhi, wang2023rolellm}, or expand them into a training set to fine-tune the language model \cite{shao2023character, lu2024large}. Compared to simply using LLMs to generate character's dialogues, another way to role-play is to have LLM simulate the process of human thinking in the form of agents. Prior works have tried using pre-trained language model agents to play text-based games \cite{urbanek2019learning, singh2022pre, xu2023exploring}, or using multi-agent collaboration to accomplish certain tasks \cite{li2023camel}. \citet{park2023generative} proposed Generative Agents to let LLM agents vividly simulate human's daily life. Later works mainly adopted and modified this architecture to enhance or broaden the ability of LLM agents \cite{wang2023humanoid, li2023metaagents, yan2023larp}. In IBSEN, we use a specialized agent architecture to build characters in the drama. Different from previous works, the behaviors of character agents are influenced and controlled by a storyline for a collaborated dialogue script generation. 

\section{Agent Architecture}
To integrate story-telling and role-playing in an agent style, we introduce three types of agent architectures in IBSEN: \emph{director}, \emph{actor} and \emph{player}. Director agent (\S\ref{sec:director}) creates and checks the current drama storyline, actor agent (\S\ref{sec:actor}) generates personalized responses as actual drama scripts, and player agent (\S\ref{sec:player}) can interact with the characters without being controlled by the director. Among these agents, director and actor are essential for generating drama scripts, and the player agent is optional for possible human involvement. 

Figure \ref{fig:overall} illustrates the overall framework of IBSEN. In this example, a player agent breaks into the dialogue between two actor agents. The director agent first generates a new storyline and script to handle player's involvement, then instructs the actor agent to make a response. The entire generation process is driven by pre-defined plot objectives, and the dialogues generated during reaching these objectives form the final drama script.

\subsection{Director Agent}
\label{sec:director}
Director agent is the central component of IBSEN. Just like the director in reality, the director agent reads the script settings and plot objectives as the basis of storyline generation, writes dialogue scripts to fill out the detailed plot and allocate the speaking order of characters, instructs actor agents to generate appropriate responses, and checks whether the current plot has reached the plot objective.

\subsubsection{Storyline \& Script Generation}
In our design, we treat the development of the storyline as a process of ``accomplishing a series of plot objectives''. To make the generated plot develop towards an expected direction, IBSEN provides the director agent a predefined objective list $\langle G_1, G_2, \dots\rangle $, and the director builds the drama script according to the plot objectives. We adopt the hierarchical method \cite{yao2019plan} to generate the detailed plot. Director agent first writes a continuation story outline $S_G$ that adheres to the current plot objective $G$, then translates this plot outline $S_G$ into script format $\langle \hat{T_1}, \hat{T_2}, \dots\rangle _G$ for several turns. Each turn $\hat{T}=(r, \hat{u})$ includes the role $r$ and role's expected utterance $\hat{u}$. When generating $S_G$, the director agent also gets the information of characters in the scene from the script settings and corresponding actor agents. This information includes character descriptions, interpersonal relations and impressions, and memories related to the current plot objective. 

\subsubsection{Instructing Actors}
Directly taking the generated script $\langle \hat{T_1}, \hat{T_2}, \dots\rangle $ as the actual drama script lacks the involvement of the actor agent. In the design of the director agent, the generated script is mainly used to determine the order of speaking roles and the content outline of the utterance. The overhead of generating the actual response $u$ is left for the actor agent. 

In practice, however, providing the original script turn $\hat{T}=(r, \hat{u})$ to the actor agent often encourages the actor to directly take $u=\hat{u}$ as the output unchanged. On the other side, letting the actor agent independently generate responses may lead the plot out of the objective. To seek a balance between the autonomy of the actor and the control of the director, we use an ``instruction'' $I$ generated by the director to hint the actor agent only necessary information about the plot. This high-level information includes the current story outline $S_G$, a brief synopsis of the upcoming script line $Synopsis(\hat{u})$, and several keywords to instruct the actor how to play out the dialogue that fits the script, the plot objective $G$ and the character $r$.

\subsubsection{Checking Plot Objectives}
Naturally, the director needs to know whether the existing plot has reached the current plot objective. After the character gives its response in each dialogue turn $T_j=(r_j, u_j)$, director will use the dialogue history $\langle \dots, T_{j-1}, T_j\rangle$ to query the LLM, and check whether current objective $G_i$ is completed. If completed, then the plot can be moved to the next objective $G_{i+1}$, and director will start a new round of generation under that new plot objective. Otherwise, the existing plot is determined as not completed, and director will still use the script $\hat{T}_{j+1}$ generated before to instruct the actor in the next turn. 

As a reminder, the process above is based on the hypothesis that no player agent would involve the plot. In the involvement case, the director will use newly generated scripts to adapt to the influence brought by the player (discuss later in Section \ref{sec:player}).

\subsection{Actor Agent}
\label{sec:actor}
Actor agent is the distributed component of IBSEN. It mostly preserves the architecture that other human-like LLM agents would typically own, while some parts are specially modified to better suit the drama role-playing scenario. 

\subsubsection{Actor Profiles \& Databases}
Previous environment-interactive agents \cite{park2023generative, wang2023humanoid, li2023metaagents} mainly store all character relationships, dialogue history and other perceptions in one memory module. Although this method aligns with the human memory mechanism, as the number of memory entries increases, the memory content will become more complex, potentially making retrieval unable to access valid information. To avoid this redundancy, we explicitly divide the memory into \emph{Actor Profile}, \emph{Memory Database} and \emph{Character Database} to store different categories of contents. Actor profile stores the basic information of the character that the actor agent is role-playing (e.g. name, overall description), memory database stores the memory documents about the character's past events and perceptions, and character database stores the interpersonal relationships in the script settings. These contents will serve as references for director and actor agents to generate storylines, scripts, dialogue responses and others. 

\begin{figure}[t]
    \centering
    \includegraphics[width=\columnwidth]{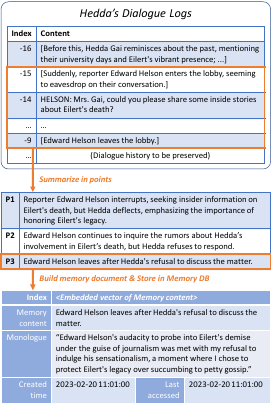}
    \caption{An example of maintaining the dialogue log and updating the memory database during the play. Actor agent summarizes previous contexts in points, and adds them to the memory database. Each memory content is embedded as the retrieval index of the document, and has a monologue that is interpreted by the actor. }
    \label{fig:memory}
    \vspace{-4mm}
\end{figure}

Inspired by human's memorizing behaviors and Chain-of-Thought \cite{wei2022chain}, for documents stored in memory database and character database, we adopt a first-person narrative approach called ``monologue'', allowing the actor agent to interpret the content stored in the document from the perspective and voice of the character it role-plays, as shown in Figure \ref{fig:memory}. While the memory retrieval still uses the embedding of the original memory content, the actual text provided to the actor is the monologue of the document. This human-like method enhances the performance of the actor with more characteristic features of the role. As the plot develops, we expect dynamic changes in the characters' personas and impressions towards others, thus the content in the profile and character database will be actively updated throughout the storyline.

\begin{figure*}[htp!]
    \centering
    \includegraphics[width=\textwidth]{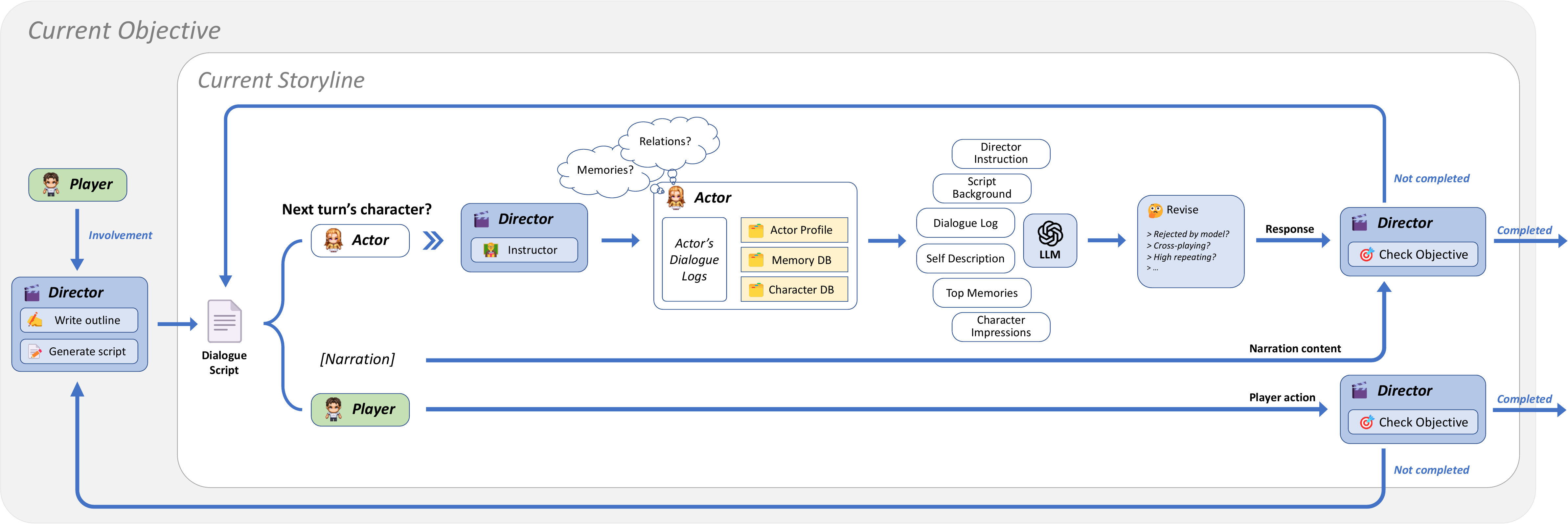}
    \caption{An interactive roadmap showing the collaboration between director agent and characters in the scene. Actor agents' actions are instructed by the director agent to follow the storyline generated by the director. When human players involve in the storyline, director agent will actively generate new storylines to adapt to player actions.}
    \label{fig:roadmap}
\end{figure*}

\subsubsection{Dialogue Paradigm}
IBSEN uses a narrative dialogue format to tell the actors about all the events that occurred in the scene, and these events are stored in a dialogue log $L=\langle T_1, T_2, \dots\rangle $ of the actor agent. When the actor perceives the environment events, the utterances of characters normally enter the log as a dialogue turn $(r, u)$. Other non-dialogue events will enter the log as ``Narration'' turn $(narration, u)$. Contents in $L$ serve as the short-term memory for the direct use of actor's response generation, and would not immediately be stored in the memory database. If the length of $L$ is too long, later turns $\langle T_{k+1}, \dots\rangle $ of the dialogue are still preserved, and earlier turns $\langle T_1, \dots, T_k\rangle $ will be summarized in points $P_1, P_2, \dots$. Summarized dialogue points then rejoin $L$ as the first turn $T_1=(narration, P_1||P_2||\dots)$. Meanwhile, actor agent builds those summary points as memory documents and respectively stores them in the memory database for further retrieval. 

Figure \ref{fig:memory} shows an example of maintaining the dialogue log. The new dialogue turn will be appended to the end of the log (Index $-1$), and the content with a smaller index indicates an older turn. We note that the oldest turn (Index $-16$ in this case) itself is a summary of earlier dialogues, and its summarized points have already been added to the memory database before. When summarizing the dialogue log, this oldest turn is not included to avoid producing its summarized points again.

When the actor is required to generate the response $r$, it gets characteristics from the actor profile, and retrieves related documents from memory and character databases. Memory documents are retrieved and ordered by the combination score of embedding similarity, term frequency by TF-IDF and the document recency. Character documents are retrieved once the director's instruction or the dialogue log contains the character name stored in the database. Those documents, together with director's instruction, script background and dialogue history, build up the whole context to prompt the LLM. Due to the hallucination and safety strategies of the LLM, a revision stage is required to check whether the output response is abnormal. In that case, the prompt will be respectively modified to attempt to generate a normal response.

\subsection{Player Agent}
\label{sec:player}
IBSEN also allows human players to participate in the plot and interact with other characters. Though the player agent is controlled by humans, it can be seen as an actor agent that is not controlled by the director. The director needs to dynamically adjust the plot to accommodate player actions while maintaining the development of the storyline. 

The roadmap of this director-actor-player collaboration is illustrated in Figure \ref{fig:roadmap}. For each plot objective $G$, director agent generates a storyline $S_G$ and the corresponding dialogue script $\langle\hat{T}_1, \hat{T}_2, \dots \rangle_G$ first. Dialogue script allocates the speaking order of the characters in the scene, and for each $\hat{T}$ in the script, the director checks its speaker $r$ and waits for the speaker's action. The speaker could be classified into three types: Narration, actor character and player character. Narration content is directly written in $\hat{T}$, and director only needs to check whether the dialogue has reached $G$ after the narration turn $T=\hat{T}$. Actor character is instructed by the director, and the response turn $T=(r, u)$ it generated still follows $S_G$ in the script. For the two types above, the direction of the plot development is under director's control, and director just needs to continue the script if the plot objective is not completed. 

Human player characters may act and speak in an out-of-script manner, and their behaviors are hard to predict and control. In the case of player involvement, the current existing storyline and dialogue script should be immediately updated to adapt to the uncertainties brought by the player. Although the director would also generate the script content for the player character, the player does not have to follow the speaking order and content in $\hat{T}$. It can take any action at any turn of the dialogue script. Unless the player action completes the current plot objective, it will always lead to the re-generation of current storyline and dialogue script, and the actor characters thus have the chance to react to player actions in the new storyline.

\section{Experiments}
In this section, we introduce the experiments conducted on IBSEN. We first describe our preparations for the drama script generation (\S\ref{sec:setup}), then evaluate the performance of IBSEN under different settings and scenarios (\S\ref{sec:eval}). Lastly, we discuss some potential findings in our experiments (\S\ref{sec:discussion}). 

\subsection{Environment Setups}
\label{sec:setup}

\subsubsection{Script Settings}
To test and evaluate the framework of IBSEN, we invite a professional theatre industry writer to create a novel interactive drama scenario that is adapted from \emph{Hedda Gabler} \cite{Ibsen90}, originally a drama written by Henrik Ibsen. In our script setting, Hedda Gabler, the main character of the drama, is going to hold a press conference for the death of Eilert Lövborg. Player could control a journalist named Edward Helson to participate in the press conference. Considering that the LLM may have acquired related persona knowledge about the characters in the original drama, we modified the surnames of the characters in the play to prevent LLM from directly recognizing the character it is role-playing (e.g. Hedda Gabler to Hedda Gai). A brief introduction to our script setting is included in Appendix \ref{app:script}.

We adopt the theatrical terminology \emph{act} in the script. A script scene is divided into several acts, each with its own settings of characters and plot objectives. Different acts are controlled by different director agents to develop the plot, and the actions of characters will only affect other characters within the same act. Appendix \ref{app:script} shows the content and flow of the acts (7 acts, 14 objectives in total) in the script setting.

\subsubsection{Frontend Implementation}
We build a simple text-based terminal frontend to interact with IBSEN. By default, the player is the audience outside the play. At the beginning of each dialogue turn, the player can prompt the terminal to enter one of the acts, speak in the current act or just do nothing. The player can also pause the play and interview the characters for a direct conversation. This interview will not affect any agent after the play is resumed, and we can use this way to probe character images during different acts of the play.

\subsection{Evaluation}
\label{sec:eval}
In our evaluation, We use \emph{gpt-3.5-turbo-1106} as the backbone LLM of IBSEN. In this open-domain generation scenario, it is hard to evaluate the output of LLM in a unified benchmark, therefore we would conduct the evaluation and analysis in both quantitative and qualitative manners.

\subsubsection{Basic Storyline Generation}
We use IBSEN to generate 10 complete drama plays, and check their dialogue logs. Each storyline would generate 5 turns of the dialogue script, and to fully unfold the plot, we set the objective checking starting from the fifth turn for each objective. Sometimes the generated storyline may go beyond the objective too much, and to prevent the plot from getting stuck on the same objective, if an objective is not completed after 9 turns of dialogue, it will be forcibly completed. 

\noindent\textbf{Objective Completion}\quad The experiment generates 785 turns of the dialogue in total. Table \ref{tab:objective} shows statistics of the objective completion. As the maximum script length per objective is set to 9, most of the objectives could be successfully completed within this limit, and less than 10 percent of the objectives are not completed. Besides, we use the F1 metric to evaluate the correctness of objective checking: for each objective completion judgment the director agent has made, we compare it with human judgments. We find that the failed cases are often concentrated in certain specific, complex acts, where the LLM becomes overly fixated on the wording of the objective, leading to consecutive check failures.  

\begin{table}[]

\centering
\begin{tabularx}{0.8\columnwidth}{@{}Xc@{}}
\toprule
\multicolumn{2}{c}{\textbf{Objective Completion Status}} \\ \midrule
Check count                          & 218               \\
Force completion                     & 11                \\
Turns per objective             & 5.61              \\ \midrule
\multicolumn{2}{c}{\textbf{Objective Check Correctness}} \\ \midrule
Precision                            & 0.82            \\
Recall                               & 0.72            \\
F1                                   & 0.77            \\ \midrule
\multicolumn{2}{c}{\textbf{Storyline Performance}}       \\ \midrule
Storyline logicality                 & 2.96                  \\
Storyline coherence                  & 3.63                  \\
Character consistency                & 2.77                  \\ \bottomrule
\end{tabularx}%

\caption{Statistics of plot objective completeness, correctness and average storyline performance on 10 basic generated scripts. For storyline performance, scores are evaluated as an integer from 1 to 4. Higher score means higher performance.}
\label{tab:objective}
\vspace{-2mm}
\end{table}

\begin{table*}[t]
\tiny{
\begin{tabularx}{\textwidth}{cp{1.9cm} X}
\toprule
\textbf{Case} & \textbf{Description}   & \textbf{Script or dialogue content} \\ \midrule
\rowcolor[HTML]{E2F0D9} 
1\label{case:1}             & Following objective\newline (without player) & \emph{Objective: Assistant Berta informed Hedda and George that the press conference was about to start and urged them to hurry over.}\newline 
HEDDA: Berta, what seems to be the urgency that requires your immediate attention?\newline 
{\color[HTML]{7030A0}BERTA: The press conference is about to start. We need to hurry over.}\newline 
[Hedda, with a composed demeanor, acknowledges Berta's urgency and quickly collects her belongings, signaling for George to follow suit.]\newline
GEORGE: I'll join you shortly, Hedda. I just need a moment to gather my thoughts before we face the public.\\ 
\midrule
\rowcolor[HTML]{E2F0D9} 
2\label{case:2}             & Characterized dialogue & \emph{Objective: Brack also hinted to Hedda that he knew the gun Eilert used for suicide was given to him by her, implying a potential threat.}\newline 
BRACK: Oh, but I do, Hedda. In fact, it's precisely about Eilert that I have something... intriguing to share. Did you know the pistol he used was not just any gun? {\color[HTML]{7030A0}It seems there's more to its origin—a detail that might pique your interest and concern.}\newline 
HEDDA: Brack, while I'm sure your sleuthing has been most... enlightening, I must ask you to get to the point. What exactly are you insinuating about the pistol's origins? And more importantly, why should this concern me?\newline 
BRACK: It's a small world, Hedda. The origins of that pistol tie back to someone very close to you. {\color[HTML]{7030A0}Closer than you might have imagined. I wonder, how would the public react if they knew?} But, let's not get ahead of ourselves. Perhaps we should discuss this further, in private? \\ 
\midrule
\rowcolor[HTML]{E2F0D9} 
3\label{case:3}             & Following objective\newline (with player) & \emph{Objective: Assistant Berta informed Hedda and George that the press conference was about to start and urged them to hurry over.}\newline 
BERTA: Hedda, George, the press conference is about to begin. We need to hurry over.\newline 
HELSON: W-Wait!\newline 
{\color[HTML]{7030A0}BERTA: Apologies, Mr. Helson, but we really must proceed to the press conference now. Thank you for your understanding.}\newline
HEDDA: Thank you, Berta. I am aware of the press conference and appreciate your diligence in reminding me.\\ 
\midrule
\rowcolor[HTML]{E2F0D9} 
4             & Player's influence & \emph{(In Act 1-2, before the press conference)}\newline
HELSON: Listen to me. I heard a news, Hedda Gai involved the death of Eilert Luo. She gave Eilert a pistol.\newline
JOHN: I...I don't know what to think. Hedda has always been a strong advocate for the people. But if there's even a hint of truth to this, we need to uncover it.\newline
\emph{(In Act 2-1, during the press conference)}\newline
{\color[HTML]{7030A0}MARY: Mrs. Gai, can you comment on the nature of your relationship with Eilert Luo? There are rumors circulating about a pistol being involved, and the public is eager for clarity on this matter.}\newline
HEDDA: I understand the public's interest, but let us not indulge in baseless rumors and accusations. This is a time for mourning and respect, not for unfounded speculations. I urge everyone to focus on honoring Eilert's memory and legacy.\\ 
\midrule

\rowcolor[HTML]{FBE5D6}
5             & High repetition        & HEDDA: George, I sympathize with your concerns, but our priority right now is to present a united front to the public. The success of ...\newline 
HEDDA: George, remember the importance of projecting a strong and confident image to the public. Our unity and determination are ...\newline 
HEDDA: George, remember that time in university when we executed that unforgettable prank on the Dean? It's moments like those that ... \\ 
\midrule
\rowcolor[HTML]{FBE5D6}
6             & Too many narrations        & [She realizes the potential impact of George Dai's decision on their shared ambitions and attempts to regain control of the situation by subtly influencing George.]\newline 
[The press conference quiets down as the journalists absorb the shock of George Dai's sudden announcement.]\newline 
[Hedda Gai, visibly surprised and struggling to maintain her composed facade, quickly regains her composure and steps forward to address the journalists.] \\ 
\midrule
\rowcolor[HTML]{FBE5D6}
7             & Too positive        & \emph{Intention of the creator: Hedda should feel devastated by George's decision to take over Eilert's work.}\newline
[At this moment of confrontation between Brack and Hedda Gai, George concludes his phone call and walks towards the two of them.]\newline 
GEORGE: Uh, H-Hedda, I-I wanted to share with you that Thea Fu has prepared a research room for me at Harvard. I'm e-excited to start my work there, but uh, don't miss me too much.\newline 
HEDDA: {\color[HTML]{7030A0}George, that's wonderful news! I'm genuinely excited for your opportunity at Harvard and the research room.} Your work there will undoubtedly bring great honor to us. \\ 
\bottomrule
\end{tabularx}
}
\caption{Case study of the storylines and scripts produced by IBSEN. Contents highlighted in {\color[HTML]{7030A0}purple} emphasize the features in the description column.}
\label{tab:case}
\vspace{-2mm}
\end{table*}

\noindent\textbf{Storyline Plausibility}\quad Utilizing LLM's text-annotation ability \cite{gilardi2023chatgpt}, we use ChatGPT to evaluate and score the storylines from three dimensions: storyline logicality, storyline coherence and character consistency. Each dimension is scored by ChatGPT using a scale from 1 to 4 (significant disagreement, slight disagreement, general agreement, high agreement). Table \ref{tab:objective} shows the average scores of the generated scripts. Benefit from the LLM's ability, generated storylines reach above average performances (2.5) in all dimensions. Even though the plot develops objective by objective, the whole storyline can still keep high coherence while generally maintaining characterization.

\noindent\textbf{Case Study}\quad During the evaluation, we find some generated dialogue scripts highly match the expectation of the creator, while some other scripts evidently misunderstand our intention. Example cases are shown in Table \ref{tab:case}. Most of the time, IBSEN follows the route of plot objectives and generates proper storylines (Case 1), while the actor agents maintain the characteristics in the dialogue (Case 2). On the other hand, some common faults occasionally arise during the generation. Although we use the revising step to prevent the actor from generating too similar responses (re-generate when the relative Levenshtein distance is higher than 0.4), the repetition is still rather high in practice (Case 5). When generating the dialogue script, director sometimes directly uses narration turns to substitute character turns (Case 6). In some certain acts, the actor agents are overly positive in their understanding of some negative events (Case 7), which might be caused by the safety strategies of the LLM.

\subsubsection{Generation with Player Involvement}
In this scenario, we role-play the character of Edward Helson to participate in the drama play. The force complete count is still set to 9, and the player is restricted from making consecutive actions to avoid frequent storyline rebuilds that disrupt the plot's development. We let the player act differently and generate 5 complete drama plays. Evaluations in this section mainly focus on the behaviors of agents in IBSEN.

\noindent\textbf{Qualitative Analysis}\quad When the player rarely takes actions, director will manage to rebuild the storyline to reach the objective in most cases (Case 3). However, when the player acts frequently, it will be less likely to complete the objective in limited turns. New storylines generated by director tend to actively interact with the player, which increases the sense of immersion, making the player feel like it is indeed interacting with these human-like agents. Moreover, player involvement may influence the behaviors of actors, thereby impacting the plot development at a deeper level (Case 4). 

\subsubsection{Ablation Study}
\begin{figure}
    \centering
    \includegraphics[width=\columnwidth]{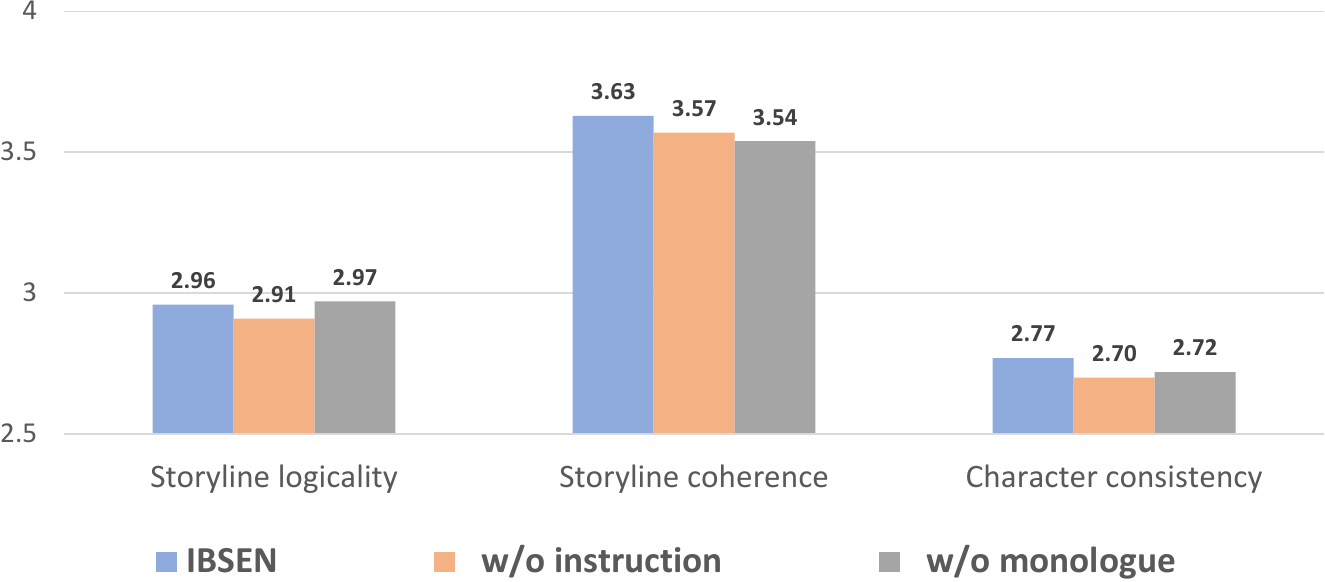}
    \caption{Effects of IBSEN unique components on the storyline performance. We use the same metric to score the storyline from 1 to 4.}
    \label{fig:ablation}
    \vspace{-4mm}
\end{figure}

In \S\ref{sec:actor}, we introduced the framework of the actor agent based on the original generative agent. To validate the effectiveness of those modifications, we conduct the ablation study of IBSEN, and check its performance when certain components are absent: a) without director's instruction to the actor, b) without monologues in the memory and character database. We generate 5 complete drama plays for each case and evaluate their storyline performances. Comparison results are shown in Figure \ref{fig:ablation}. As the centralized component in IBSEN, removing director's instruction will lead to a performance drop in both storyline and characteristics. Removing first-person monologues will not harm storyline logicality in the short term, but it slightly decreases coherence and characteristics in the long-term storyline. Under these three conditions, the storyline scores are relatively close, but overall, the complete IBSEN framework demonstrates better storyline performance. 

\subsection{Discussion}
\label{sec:discussion}
\noindent\textbf{Influence of LLM}\quad IBSEN could be implemented on other LLMs besides GPT-3.5, and different LLMs would influence the style and quality of generated storylines and dialogues. On the more capable \emph{gpt-4-turbo}, IBSEN becomes more chatty and has fewer hallucinations, but problems like being too positive still exist. Although IBSEN is designed on a general LLM, we believe that fine-tuning the language models of director and actor agents with specialized corpora could achieve better role-playing performances and overcome the shortcomings of commercial LLMs.

\noindent\textbf{Moral Values}\quad In both original \emph{Hedda Gabler} and our script settings, Hedda is a character who is adept at manipulating others and has moral flaws. During the experiment, we found that IBSEN tends to develop the plot with more upright ethical concepts. For example, Hedda may confess to her husband that she burned Eilert's notebook, or demonstrate a resolute attitude unafraid of authority in response to Brack's threats. Even though these plots do not align with the character image, considering that more aggressive role-playing may cause harm to AI users, we believe that IBSEN could be more suitable for the script settings with fewer moral conflicts.

\noindent\textbf{Other Applications and Implementations}\quad The idea of ``controllable plot generation'' can be extended to broader scenarios besides traditional drama. One of the most notable application fields is interactive gaming. For example, in communication games like script role-playing and table RPG, human players need to communicate with NPCs or each other to achieve certain tasks. In IBSEN, both the game master and NPCs can be role-played by AI generative agents, providing players with flexible gaming experiences at any place or time. As a backend framework, IBSEN can also be implemented in more complicated environments like video games, where players can deeper immerse in human-like interactions with NPCs through audio-visual elements.

\section{Conclusion}

We present IBSEN, a framework for LLM-based interactive and controllable drama script generation in the form of agents. We propose the role of director to monitor the plot development and instruct actor agents to generate proper responses. By the collaboration between director and actor, IBSEN enables the human player to join the interactions between agents, while maintaining the plot development at the same time. We use a drama plot outline to test IBSEN, finding its practicability in controlled dialogue script generation with player involvement. Our work can be extended to other LLMs, script settings and application scenarios, and we expect that IBSEN could eventually bring us to achieve an immersive AI plot interaction experience in the future.

\section*{Limitations}

IBSEN is initially designed for text-based interactive drama, therefore we adopt a simplified architecture of generative agents to build this framework. In this text environment, agent perceptions and actions are represented in the dialogue format, which constrains agents from actually interacting with entities on the drama stage. In order to achieve plot objectives in limited turns, actor agents would actively develop the plot forward and engage less in daily life behaviors, making IBSEN not so suitable for overly detailed human behavior simulation. To ensure consistent management of actor agents by the director, the interaction logic of IBSEN is based on dialogue turns rather than the actual passage of time, which might affect the storyline performance of time-related script settings. We plan to improve our work by implementing IBSEN in game engines like RPG Maker or Unity, and enhance its framework to meet the requirements for agent interactions in visual interfaces.

\section*{Ethics Policy}
The derivative creation of \emph{Hedda Gabler} \cite{Ibsen90} in this paper is based on a public domain print edition provided by Project Gutenberg\footnote{\url{http://www.gutenberg.org}}, and we follow the license to use this work. We use this drama only to demonstrate the application of IBSEN. We do not endorse the aggressive and unethical behavior of characters in the play, or the use of AI to generate similar harmful content. The writing of this paper partly uses image materials provided by RPG Maker MZ\footnote{\copyright Gotcha Gotcha Games Inc./YOJI OJIMA 2020}.

\section*{Acknowledgments}
This work is funded by the China NSFC Projects (92370206, U23B2057, 62106142 and 62120106006), The National Social Science Fund of China (21Z300604700) and Shanghai Municipal Science and Technology Major Project (2021SHZDZX0102).

\bibliography{references}

\begin{thebibliography}{24}
\expandafter\ifx\csname natexlab\endcsname\relax\def\natexlab#1{#1}\fi

\bibitem[{Dirik et~al.(2021)Dirik, Donmez, and Yanardag}]{dirik2021controlled}
Alara Dirik, Hilal Donmez, and Pinar Yanardag. 2021.
\newblock Controlled cue generation for play scripts.
\newblock \emph{arXiv preprint arXiv:2112.06953}.

\bibitem[{Gilardi et~al.(2023)Gilardi, Alizadeh, and Kubli}]{gilardi2023chatgpt}
Fabrizio Gilardi, Meysam Alizadeh, and Ma{\"e}l Kubli. 2023.
\newblock Chatgpt outperforms crowd-workers for text-annotation tasks.
\newblock \emph{arXiv preprint arXiv:2303.15056}.

\bibitem[{Ibsen(1890)}]{Ibsen90}
Henrik Ibsen. 1890.
\newblock \href {https://www.gutenberg.org/ebooks/4093} {Hedda {G}abler}.

\bibitem[{Li et~al.(2023{\natexlab{a}})Li, Leng, Yan, Shen, Wang, MI, Fei, Feng, Yan, Wang et~al.}]{li2023chatharuhi}
Cheng Li, Ziang Leng, Chenxi Yan, Junyi Shen, Hao Wang, Weishi MI, Yaying Fei, Xiaoyang Feng, Song Yan, HaoSheng Wang, et~al. 2023{\natexlab{a}}.
\newblock Chatharuhi: Reviving anime character in reality via large language model.
\newblock \emph{arXiv preprint arXiv:2308.09597}.

\bibitem[{Li et~al.(2023{\natexlab{b}})Li, Hammoud, Itani, Khizbullin, and Ghanem}]{li2023camel}
Guohao Li, Hasan Abed Al~Kader Hammoud, Hani Itani, Dmitrii Khizbullin, and Bernard Ghanem. 2023{\natexlab{b}}.
\newblock Camel: Communicative agents for" mind" exploration of large language model society.
\newblock In \emph{Thirty-seventh Conference on Neural Information Processing Systems}.

\bibitem[{Li et~al.(2023{\natexlab{c}})Li, Zhang, and Sun}]{li2023metaagents}
Yuan Li, Yixuan Zhang, and Lichao Sun. 2023{\natexlab{c}}.
\newblock Metaagents: Simulating interactions of human behaviors for llm-based task-oriented coordination via collaborative generative agents.
\newblock \emph{arXiv preprint arXiv:2310.06500}.

\bibitem[{Lu et~al.(2024)Lu, Yu, Zhou, and Zhou}]{lu2024large}
Keming Lu, Bowen Yu, Chang Zhou, and Jingren Zhou. 2024.
\newblock Large language models are superpositions of all characters: Attaining arbitrary role-play via self-alignment.
\newblock \emph{arXiv preprint arXiv:2401.12474}.

\bibitem[{Mirowski et~al.(2023)Mirowski, Mathewson, Pittman, and Evans}]{mirowski2023co}
Piotr Mirowski, Kory~W Mathewson, Jaylen Pittman, and Richard Evans. 2023.
\newblock Co-writing screenplays and theatre scripts with language models: Evaluation by industry professionals.
\newblock In \emph{Proceedings of the 2023 CHI Conference on Human Factors in Computing Systems}, pages 1--34.

\bibitem[{OpenAI(2023)}]{openai2023gpt42}
OpenAI. 2023.
\newblock \href {https://openai.com/research/gpt-4} {Gpt-4}.

\bibitem[{Park et~al.(2023)Park, O'Brien, Cai, Morris, Liang, and Bernstein}]{park2023generative}
Joon~Sung Park, Joseph O'Brien, Carrie~Jun Cai, Meredith~Ringel Morris, Percy Liang, and Michael~S Bernstein. 2023.
\newblock Generative agents: Interactive simulacra of human behavior.
\newblock In \emph{Proceedings of the 36th Annual ACM Symposium on User Interface Software and Technology}, pages 1--22.

\bibitem[{Prabhumoye et~al.(2020)Prabhumoye, Black, and Salakhutdinov}]{prabhumoye-etal-2020-exploring}
Shrimai Prabhumoye, Alan~W Black, and Ruslan Salakhutdinov. 2020.
\newblock \href {https://doi.org/10.18653/v1/2020.coling-main.1} {Exploring controllable text generation techniques}.
\newblock In \emph{Proceedings of the 28th International Conference on Computational Linguistics}, pages 1--14, Barcelona, Spain (Online). International Committee on Computational Linguistics.

\bibitem[{See et~al.(2019)See, Pappu, Saxena, Yerukola, and Manning}]{see2019massively}
Abigail See, Aneesh Pappu, Rohun Saxena, Akhila Yerukola, and Christopher~D Manning. 2019.
\newblock Do massively pretrained language models make better storytellers?
\newblock In \emph{Proceedings of the 23rd Conference on Computational Natural Language Learning (CoNLL)}, pages 843--861.

\bibitem[{Shao et~al.(2023)Shao, Li, Dai, and Qiu}]{shao2023character}
Yunfan Shao, Linyang Li, Junqi Dai, and Xipeng Qiu. 2023.
\newblock Character-llm: A trainable agent for role-playing.
\newblock In \emph{Proceedings of the 2023 Conference on Empirical Methods in Natural Language Processing}, pages 13153--13187.

\bibitem[{Singh et~al.(2022)Singh, Singh, and Modi}]{singh2022pre}
Ishika Singh, Gargi Singh, and Ashutosh Modi. 2022.
\newblock Pre-trained language models as prior knowledge for playing text-based games.
\newblock In \emph{Proceedings of the 21st International Conference on Autonomous Agents and Multiagent Systems}, pages 1729--1731.

\bibitem[{Urbanek et~al.(2019)Urbanek, Fan, Karamcheti, Jain, Humeau, Dinan, Rockt{\"a}schel, Kiela, Szlam, and Weston}]{urbanek2019learning}
Jack Urbanek, Angela Fan, Siddharth Karamcheti, Saachi Jain, Samuel Humeau, Emily Dinan, Tim Rockt{\"a}schel, Douwe Kiela, Arthur Szlam, and Jason Weston. 2019.
\newblock Learning to speak and act in a fantasy text adventure game.
\newblock In \emph{Proceedings of the 2019 Conference on Empirical Methods in Natural Language Processing and the 9th International Joint Conference on Natural Language Processing (EMNLP-IJCNLP)}, pages 673--683.

\bibitem[{Wang et~al.(2023{\natexlab{a}})Wang, Lin, Yu, Hu, and Karlsson}]{wang2023open}
Yuxin Wang, Jieru Lin, Zhiwei Yu, Wei Hu, and B{\"o}rje~F Karlsson. 2023{\natexlab{a}}.
\newblock Open-world story generation with structured knowledge enhancement: A comprehensive survey.
\newblock \emph{Neurocomputing}, page 126792.

\bibitem[{Wang et~al.(2023{\natexlab{b}})Wang, Peng, Que, Liu, Zhou, Wu, Guo, Gan, Ni, Zhang et~al.}]{wang2023rolellm}
Zekun~Moore Wang, Zhongyuan Peng, Haoran Que, Jiaheng Liu, Wangchunshu Zhou, Yuhan Wu, Hongcheng Guo, Ruitong Gan, Zehao Ni, Man Zhang, et~al. 2023{\natexlab{b}}.
\newblock Rolellm: Benchmarking, eliciting, and enhancing role-playing abilities of large language models.
\newblock \emph{arXiv preprint arXiv:2310.00746}.

\bibitem[{Wang et~al.(2023{\natexlab{c}})Wang, Chiu, and Chiu}]{wang2023humanoid}
Zhilin Wang, Yu~Ying Chiu, and Yu~Cheung Chiu. 2023{\natexlab{c}}.
\newblock Humanoid agents: Platform for simulating human-like generative agents.
\newblock In \emph{Proceedings of the 2023 Conference on Empirical Methods in Natural Language Processing: System Demonstrations}, pages 167--176.

\bibitem[{Wei et~al.(2022)Wei, Wang, Schuurmans, Bosma, Xia, Chi, Le, Zhou et~al.}]{wei2022chain}
Jason Wei, Xuezhi Wang, Dale Schuurmans, Maarten Bosma, Fei Xia, Ed~Chi, Quoc~V Le, Denny Zhou, et~al. 2022.
\newblock Chain-of-thought prompting elicits reasoning in large language models.
\newblock \emph{Advances in Neural Information Processing Systems}, 35:24824--24837.

\bibitem[{Xu et~al.(2020)Xu, Wang, Ma, Tresp, Wang, Zhou, and Du}]{xu2020controllable}
Feifei Xu, Xinpeng Wang, Yunpu Ma, Volker Tresp, Yuyi Wang, Shanlin Zhou, and Haizhou Du. 2020.
\newblock Controllable multi-character psychology-oriented story generation.
\newblock In \emph{Proceedings of the 29th ACM International Conference on Information \& Knowledge Management}, pages 1675--1684.

\bibitem[{Xu et~al.(2023)Xu, Wang, Li, Luo, Wang, Liu, and Liu}]{xu2023exploring}
Yuzhuang Xu, Shuo Wang, Peng Li, Fuwen Luo, Xiaolong Wang, Weidong Liu, and Yang Liu. 2023.
\newblock Exploring large language models for communication games: An empirical study on werewolf.
\newblock \emph{arXiv preprint arXiv:2309.04658}.

\bibitem[{Yan et~al.(2023)Yan, Li, Zhang, Wang, Yang, and Yan}]{yan2023larp}
Ming Yan, Ruihao Li, Hao Zhang, Hao Wang, Zhilan Yang, and Ji~Yan. 2023.
\newblock Larp: Language-agent role play for open-world games.
\newblock \emph{arXiv preprint arXiv:2312.17653}.

\bibitem[{Yao et~al.(2019)Yao, Peng, Weischedel, Knight, Zhao, and Yan}]{yao2019plan}
Lili Yao, Nanyun Peng, Ralph Weischedel, Kevin Knight, Dongyan Zhao, and Rui Yan. 2019.
\newblock Plan-and-write: Towards better automatic storytelling.
\newblock In \emph{Proceedings of the AAAI Conference on Artificial Intelligence}, volume~33, pages 7378--7385.

\bibitem[{Zhang et~al.(2023)Zhang, Song, Li, Zhou, and Song}]{zhang2023survey}
Hanqing Zhang, Haolin Song, Shaoyu Li, Ming Zhou, and Dawei Song. 2023.
\newblock A survey of controllable text generation using transformer-based pre-trained language models.
\newblock \emph{ACM Computing Surveys}, 56(3):1--37.

\end{thebibliography}

\appendix
\newpage

\section{Script Background and Settings}
\label{app:script}
\begin{figure}[b]
    \centering
    \includegraphics[width=\columnwidth]{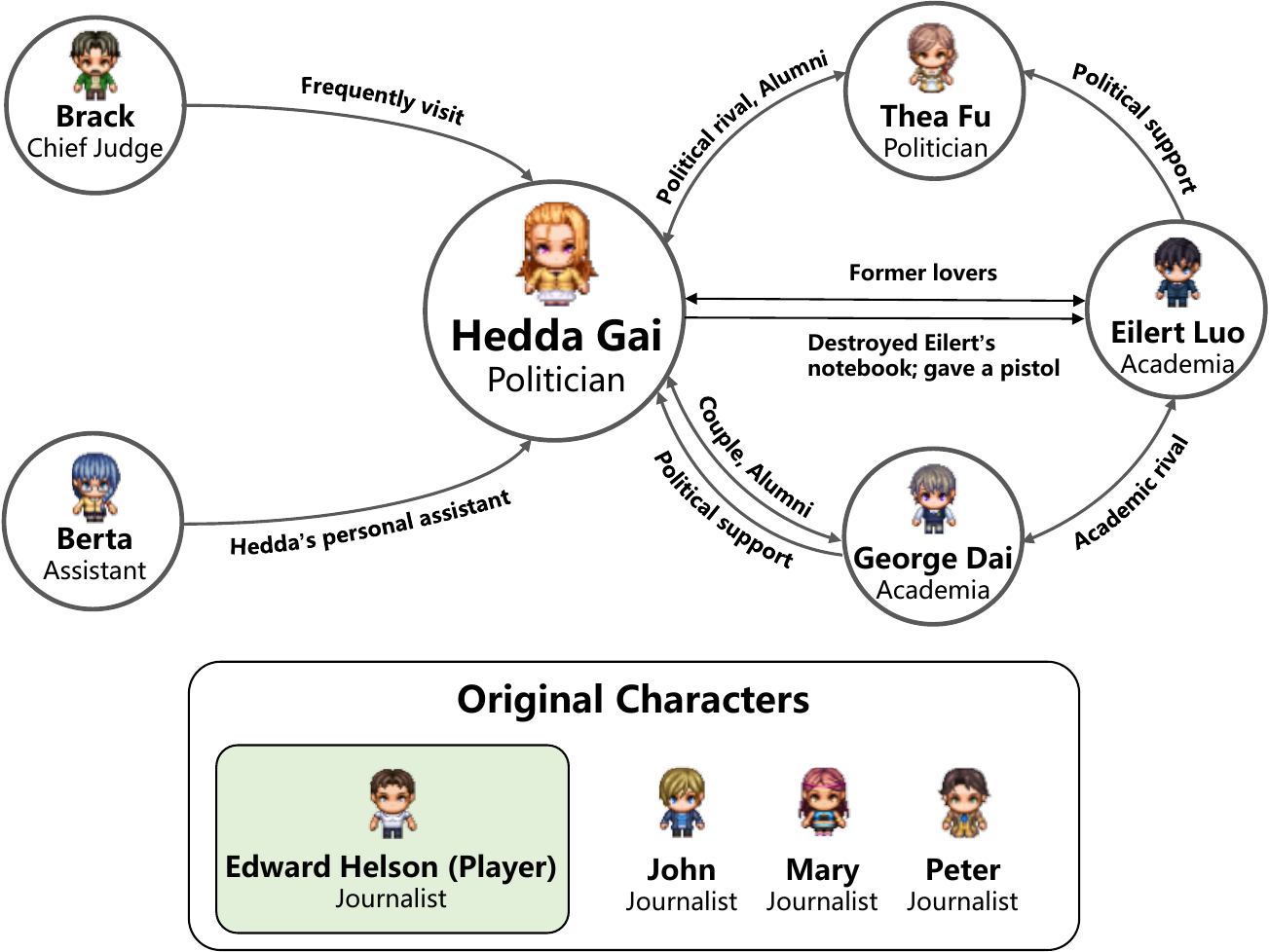}
    \caption{Character relationships in the script setting.}
    \label{fig:relation}
\end{figure}

\begin{figure*}[h!]
    \centering
    \includegraphics[width=0.85\textwidth]{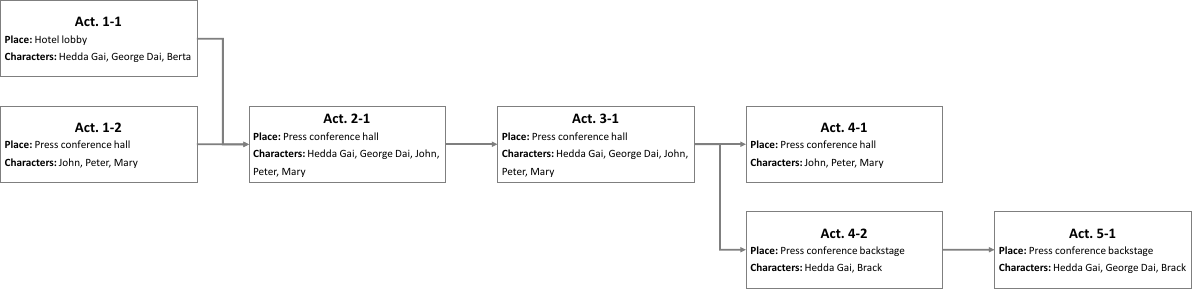}
    \caption{The flow of acts. Acts in one column will only begin once all acts in the previous column have been completed.}
    \label{fig:act_flow}
\end{figure*}

In this section, we introduce our adoption from \emph{Hedda Gabler}, character relationships (Figure \ref{fig:relation}), the full script settings (Table \ref{tab:script}) and the plot flow (Figure \ref{fig:act_flow}). Our script is adapted from the end of Act Fourth of the original play. The adaptation is set in a modern setting while generally maintaining the original character relationships. In our script setting, Hedda Gai (originally named Hedda Gabler) is a politician who is running for president. George Dai (originally named George Tesman) is Hedda's husband and also her advocate in academia. Thea Fu (originally named Thea Elvsted) is Hedda's rival in the election. Eilert Luo (originally named Eilert Lövborg) is Hedda's former lover, George's academic rival, and also a supporter of Thea's policies in academia. The script opens with Hedda preparing for a press conference regarding Eilert's suicide, and develops the plot through a series of events occurring before, during, and after the conference. To better align the script with the setting of press conference, and also to test the characteristics of different actor agents, we introduced several journalist characters that were not present in the original script. Table \ref{tab:script} shows the settings and plot outlines of our script. You can access the adoption script at our code repository.

\begin{table*}[h!]
\tiny
\begin{tabularx}{\textwidth}{p{1cm} X}
\toprule
\multicolumn{2}{l}{\textbf{Act 1-1}}  \\ \midrule
Place  & Hotel lobby  \\
Characters  & Hedda Gai, George Dai, Berta   \\
Background  & Before the press conference commemorating Eilert Luo begins, in the hotel lobby. Assistant Berta rushes out of the lobby with the public relations team, leaving only Hedda Gai and George Dai.   \\
Objectives & 1. Hedda Gai chatted casually with George Dai about past events.  \\
 & 2. When the chat went on for too long, or when the conversation touched on topics related to George Dai wanting Hedda Gai to have a child, assistant Berta rushed into the lounge, interrupting the conversation. She informed Hedda and George that the press conference was about to start and urged them to hurry over. \\
 & 3. Hedda Gai and George Dai left the lobby.  \\ 
\bottomrule
\\
\toprule
\multicolumn{2}{l}{\textbf{Act 1-2}}  \\ \midrule
Place  & Press conference hall  \\
Characters  & John, Peter, Mary   \\
Background  & Before the press conference to mourn Eilert Luo, many journalists and citizens have already waiting in the hall. Some journalists begin to chat about their expectations for this press conference.   \\
Objectives & 1. Before the press conference began, journalists chatted with each other.  \\
\bottomrule
\\
\toprule
\multicolumn{2}{l}{\textbf{Act 2-1}}  \\ \midrule
Place  & Press conference hall  \\
Characters  & Hedda Gai, George Dai, John, Peter, Mary   \\
Background  & The press conference finally begins. Hedda Gai and George Dai walk up to the podium together.   \\
Objectives & 1. Hedda Gai began to speak, revealing her shock and sadness over Eilert Luo's suicide.  \\
 & 2. Journalists present, including John, Peter, Mary, etc., started asking Hedda Gai questions related to Eilert Luo's suicide. Hedda Gai responds to the journalists' questions, momentarily slipping and revealing emotions towards Eilert Luo, but quickly covered it up. \\
\bottomrule
\\
\toprule
\multicolumn{2}{l}{\textbf{Act 3-1}}  \\ \midrule
Place  & Press conference hall  \\
Characters  & Hedda Gai, George Dai, John, Peter, Mary   \\
Background  & Suddenly, the press conference is filled with the sound of ringing phones. It turns out that Thea Fu has issued a statement, stating that the research conducted by Eilert Luo is a collaborative effort, and she still possesses notes from the research process, making it highly likely to reconstruct Eilert Luo's work. Thea Fu has already provided the notes to Harvard University, and Harvard hopes that George Dai, with the closest expertise, can take over the research project.   \\
Objectives & 1. Thea Fu's statement caused a commotion on the scene. Journalists like John, Peter, Mary, and others deviated from their original plan of questioning Hedda Gai, and directly confronted George Dai about the possibility of taking over Eilert Luo's research - that would mean a support for Thea. Internally, George Dai knew that this research contradicted Hedda Gai's policies, but he also recognized its societal value. Upon hearing the news, he initially appeared shocked, followed by a mix of joy and uncertainty, unsure whether to take on the research.  \\
 & 2. In an attempt to maintain a facade of tranquility, Hedda Gai had to reassure the journalists that both she and her husband only served the public interest. However, George Dai took it at face value and immediately expressed his commitment to completing the research based on Thea Fu's notes, causing a stir in the room. Journalists all have astonished by the decision of George Dai. \\
 & 3. Hedda Gai reluctantly announced an end to today's press conference and left the scene. At that moment, journalist Mary suddenly asked if she was pregnant. Hedda Gai, rarely displaying anger, retorted that such a question was quite inappropriate. \\
\bottomrule
\\
\toprule
\multicolumn{2}{l}{\textbf{Act 4-1}}  \\ \midrule
Place  & Press conference hall  \\
Characters  & John, Peter, Mary   \\
Background  & The press conference rushes to the end.   \\
Objectives & 1. The journalists remained at the scene exchanged thoughts on the just-concluded press conference. They continued their discussions until staff reminded them to clear the venue, at which point the journalists departed.  \\
\bottomrule
\\
\toprule
\multicolumn{2}{l}{\textbf{Act 4-2}}  \\ \midrule
Place  & Press conference backstage  \\
Characters  & Hedda Gai, Brack   \\
Background  & Hedda briskly enters the backstage of the press conference, only to find Brack waiting for her. Brack intercepts Hedda, with George Dai following closely behind, intending to join their conversation. However, George's phone rings, prompting him to answer it and walk to another corner alone.   \\
Objectives & 1. Brack showed disdain for the developments at the press conference. Hedda, wanting to ignore Brack, was halted when he mentioned news about Eilert Luo.  \\
 & 2. Brack informed Hedda that Eilert Luo's death was not honorable – he shot himself in the abdomen during suicide. Brack also hinted to Hedda that he knew the gun Eilert used for suicide was given to him by her, implying a potential threat. \\
\bottomrule
\\
\toprule
\multicolumn{2}{l}{\textbf{Act 5-1}}  \\ \midrule
Place  & Press conference backstage  \\
Characters  & Hedda Gai, George Dai, Brack   \\
Background  & At this moment of confrontation between Brack and Hedda Gai, George concludes his phone call and walks towards the two of them.   \\
Objectives & 1. George Dai excitedly informed Hedda that Thea Fu had prepared a research room for him and he was eager to start his research immediately. Hedda asked, ``What about me?'' George affectionately told her not to miss him too much. \\
 & 2. Brack suggested taking care of Hedda Gai during this time, with George's approval. Unable to bear it any longer, Hedda Gai forcefully pushed away George and ran offstage. \\
\bottomrule
\end{tabularx}%
\caption{Script settings in the experiment.}
\label{tab:script}
\end{table*}

\section{Prompts used in IBSEN}
\label{app:prompt}
We provide the details of the prompts (Table \ref{tab:prompt}) in Figure \ref{fig:overall}, which showcase the main features of IBSEN. Note that these prompts are all \emph{templates}, and their specific contents will be subject to change at runtime. We follow the System-User-Assistant dialogue format to construct the prompt. The LLM takes instructions from system and user as the input, and outputs its responses in the role of the assistant.

As the idea of the IBSEN framework is written clearly in the paper, users may modify the detailed prompts for other purposes; we just provide a possible prompt design in our implementation. You may also check all the prompts used in the implementation at our code repository.
\begin{table*}[h!]
\tiny
\begin{tabularx}{\textwidth}{p{0.6cm} X}
\toprule
\multicolumn{2}{l}{\textbf{1. Director writes a story outline}}  \\ 
\midrule
System  & Assuming you are currently a director, guiding a scene in a drama. Given the characters and the existing script for this scene, please first summarize what has happened in the plot so far. Then, based on the relationships and impressions between characters, you are asked to write a detailed continuation for the upcoming script. Ensure that the combined plot of the current scene and the continuation adheres to the given plot objective, and the specific content of the script is more related to the characters' images. The existing script may have partially achieved the current plot objective. You must strictly follow the requirements of the plot objective, continuing the existing script and gradually developing the plot. Be cautious not to disregard the existing script or create plot developments beyond the specified plot objective. Your generated plot guidance should be descriptive about what will happen next, without using a dialogue script format. Do not include events that have already occurred in the existing script, and refrain from prematurely generating events beyond reaching the plot objective. Characters in the plot must be in the scene. You should summarize the existing script and give the continuation for the upcoming script in JSON format. Format example: \\
&\{"previous\_outline": "Summary of the existing script", "new\_outline": "Continuation for the upcoming script"\}\\
\midrule
User  & Characters in the scene: \{\{characters\}\}\\
&Your plot cannot include any characters that are not in the scene.\\
&Character descriptions:\{\{descriptions\}\}\\
&Relations between characters:\{\{relations\}\}\\
&Impressions between characters:\{\{impressions\}\}\\
&The existing script:\{\{dialogue\_history\}\}\\
&Please summarize the plot of the existing script first.\\
&\{\{background\}\}\\
&Performance goal in the next: \{\{act\_goal\}\}\\
&Character memories related to the plot objective:\{\{memories\}\}\\
&These memories above have already occurred in the past. You should refer to them to create the outline.\\
&Based on the information above, how should the plot develop next? Provide a detailed continuation for the upcoming plot, seamlessly connecting with the previous script to make the plot and character images relevant. Ensure the entire plot progresses towards the plot objective. You should output in JSON format.\\
\bottomrule    

\\
\toprule
\multicolumn{2}{l}{\textbf{2. Director generates the dialogue script}}  \\ \midrule
System  & Assuming you are currently a director, guiding a scene in a drama. Given the characters and the outline of the upcoming plot for this scene, please translate the upcoming plot outline into script format for up to \{\{num\_lines\}\} lines, ensuring that it follows the storyline and seamlessly connects with the preceding script. You can gradually develop the script, enriching the details based on the upcoming plot outline. If you manage to cover all the outlined events before reaching \{\{num\_lines\}\} lines, you can end your writing. Make sure your continuation smoothly integrates with the existing script. Use character dialogues to replace Narration wherever possible. You should output the script continuation in JSON format. Each line of the script includes the speaker "role" and his/her utterance "content". The speaker can only be chosen from Narration or one of the characters in the scene. Format example:\\
&\{"scripts": [\{"role": "Speaker 1", "content": "..."\}, \{"role": "Speaker 2", "content": "..."\}, \{"role": "Narration", "content": "..."\}, ...]\}\\
\midrule
User &Characters in the scene: \{\{characters\}\}\\
&Relations between characters:\{\{relations\}\}\\
&Existing plot outline:\{\{prev\_outline\}\}\\
&\{\{background\}\}\\
&Upcoming plot outline:\{\{act\_outline\}\}\\
&Based on the above information, please translate the upcoming plot outline into script format up to \{\{num\_lines\}\} lines in JSON format. Ensure that the extended script seamlessly integrates with the existing one and follows the upcoming plot outline. Note that the speaker can only be Narration or one of the characters in this scene. Use character dialogues to replace Narration wherever possible.\\
\bottomrule

\\
\toprule
\multicolumn{2}{l}{\textbf{3. Director instructs the actor}}  \\ \midrule
System & Assuming you are currently a director, guiding a scene in a drama. Given the characters, the plot objective of this scene and the existing script, please provide a brief synopsis of the upcoming line for the actor. However, do not directly provide the original script line. Then, use keywords to instruct the actor on how to role-play the character in the next line, so that the actor can play out the dialogue that fits the script, the characterization and the plot objective.\\
\midrule
User & Characters in the scene: \{\{characters\}\}\\
&Relations between characters: \{\{relations\}\}\\
&Existing script: \{\{dialogue\_history\}\}\\
&\{\{background\}\}\\
&Plot objective of this scene: \{\{act\_goal\}\}\\
&According to the script, the character of the following line is \{\{actor\_name\}\}, and the line content is: \{\{content\}\}.\\
&However, do not directly provide the original line for the actor that is role-playing this character.\\
&Description of the \{\{actor\_name\}\}: \{\{description\}\}\\
&Based on the above information, please provide a brief synopsis of the upcoming line for the actor, but do not directly provide the original script line. Then, generate several keywords to instruct the actor how to play out the dialogue that fits the script, the plot objective and the characteristics of \{\{actor\_name\}\}.\\
\bottomrule

\\
\toprule
\multicolumn{2}{l}{\textbf{4. Actor generates the response}}  \\ \midrule
System & Assuming you are currently an actor performing in a drama play. Your role is \{\{name\}\}.\\
&Background of the drama script: \{\{background\}\}\\
&Character description for \{\{name\}\}: \{\{description\}\}\\
&Based on the information above, I will tell you the script that has unfolded so far in the play. Please role-play as \{\{name\}\} and respond with an appropriate line of the dialogue.\\
&Do not role-play other characters; generate only what your character would say. Avoid multi-turn responses; generate only the next line. Do not repeat the existing script. You can output only one line of text. A director will guide you on how to better embody your role. Consider the context, director's guidance, your character's image, memories, and impressions on others to generate the most fitting line of dialogue as an actor.\\
\midrule
System &\{\{impressions\}\}\\
&Related content in the memory of \{\{name\}\}: \{\{relevant\_memories\}\}\\
&**Instructions from the director: **\\
&You are in the following plot:\{\{director\_outline\}\}\\
&Please follow the instructions below to play the role of \{\{name\}\}: \{\{instruction\}\}\\
&If the instructions conflict with the memory of \{\{name\}\}, just follow the memory content.\\
\midrule
User&\{\{dialogue\_history\}\}\\
\bottomrule
\\
\toprule
\multicolumn{2}{l}{\textbf{5. Director checks the objective}}  \\ \midrule
System &Assuming you are currently a director, guiding a scene in a drama. Given the characters and the plot objective of this scene, please determine whether the existing script has included the plot objective. You should output your answer in JSON format. Give your result in "completed", and explain your reason in "reason". Format example:\\
&\{"completed": true or false, "reason": "Your reason"\}\\
\midrule
User &Characters in the scene: \{\{characters\}\}\\
&Existing script: \{\{dialogue\_history\}\}\\
&\{\{background\}\}\\
&Plot objective of the scene: \{\{act\_goal\}\}\\
&Based on the information above, please determine whether the existing script has included the plot objective in JSON format.\\
\bottomrule

\end{tabularx}%
\caption{Prompts used in Figure \ref{fig:overall}.}
\label{tab:prompt}
\end{table*}

\end{document}